# "Be My Cheese?": Assessing Cultural Nuance in Multilingual LLM Translations

Madison Van Doren, Cory Holland
Appen

## 1. Abstract

This pilot study explores the localisation capabilities of state-of-the-art multilingual AI models when translating figurative language, such as idioms and puns, from English into a diverse range of global languages. It expands on existing LLM translation research and industry benchmarks, which emphasise grammatical accuracy and token-level correctness, by focusing on cultural appropriateness and overall localisation quality—critical factors for real-world applications like marketing and e-commerce.

To investigate these challenges, this project evaluated a sample of 87 LLM-generated translations of e-commerce marketing emails across 24 regional dialects of 20 languages. Human reviewers fluent in each target language provided quantitative ratings and qualitative feedback on faithfulness to the original's tone, meaning, and intended audience. Findings suggest that, while leading models generally produce grammatically correct translations, culturally nuanced language remains a clear area for improvement, often requiring substantial human refinement. Notably, even high-resource global languages, despite topping industry benchmark leaderboards, frequently mistranslated figurative expressions and wordplay.

This work challenges the assumption that data volume is the most reliable predictor of machine translation quality and introduces cultural appropriateness as a key determinant of multilingual LLM performance—an area currently underexplored in existing academic and industry benchmarks. As a proof of concept, this pilot highlights limitations of current multilingual AI systems for real-world localisation use cases. Results of this pilot support the opportunity for expanded research at greater scale to deliver generalisable insights and inform deployment of reliable machine translation workflows in culturally diverse contexts.

## 2. Introduction

As multilingual large language models (LLMs) are increasingly integrated into global content workflows, understanding their ability to produce culturally appropriate translations is critical for effective localisation. Cultural nuance, especially figurative language such as wordplay, is central to effective human communication.

This pilot study investigates how well state-of-the-art multilingual LLMs handle figurative language when translating from English into a diverse range of global languages, addressing three core research questions: (1) How reliably do LLMs translate idiomatic and figurative language across languages with varying resource availability and linguistic characteristics? (2) Does linguistic proximity or shared structural features with English predict translation success? (3) Are culturally nuanced translations more accurate in highly resourced, globally distributed languages than in smaller, regional ones?

Through a systematic evaluation of LLM-generated translations across 24 dialects of 20 global languages, this research provides insight into current model limitations, cross-linguistic patterns, and the ongoing role of human expertise in localisation workflows.

## 3. Related Work

Recent advances in large language models (LLMs) demonstrate significant progress in the multilingual translation capabilities of generative AI. Foundational work by Mujadia et al. (2023) provides a comprehensive assessment of LLM translation capabilities between English and 22 Indian languages, revealing notable disparities in performance across high-resource and low-resource settings. Their findings underscore the importance of in-context learning, which substantially improves translation quality for underrepresented regional dialects. Similarly, Hu et al. (2024) introduce GenTranslate, a generative framework that improves multilingual speech and text translation on standard benchmarks, particularly for low-resource languages, by leveraging LLMs' context awareness and reasoning abilities. This work supports the notion that multilingual LLM performance is improved by in-context learning.

Beyond translation accuracy, scholars have increasingly scrutinised the cultural and ethical implications of LLM-generated outputs. AlKhamissi et al. (2024) explore cultural alignment across languages and regions, demonstrating that LLMs more accurately reflect cultural knowledge when prompted in a region's dominant language. However, they also identify persistent representation disparities, particularly for historically marginalised cultures. Building on this, Li et al. (2024) propose CultureLLM, a framework that incorporates cultural context through multilingual and culturally diverse data. Their results show measurable gains in cultural appropriateness, though challenges remain in low-resource settings. These findings align with broader concerns about linguistic equity and the accurate representation of diverse cultural perspectives in [LLM training data](#).

Another critical dimension of multilingual LLM research is fairness and bias. Zhao et al. (2024) conduct a multilingual analysis of gender bias, identifying significant cross-linguistic variation in how stereotypes manifest in LLM outputs. They assert that models trained primarily on English data tend to exhibit greater bias when generating responses in non-English languages. Complementing this work, Sterlie et al. (2024) extend traditional non-discrimination criteria to generative models. By applying these fairness metrics to gendered language tasks, they demonstrate systematic occupational and descriptive biases and propose reformulations that improve fairness in generative outputs.

Finally, questions of safety and robustness continue to motivate LLM evaluation research. Prior work by the present author (Van Doren, 2025) outlines principles for building responsible and reliable AI systems, with a focus on transparent evaluation frameworks that integrate human alignment and adversarial testing to mitigate potential personal, societal, and legal risks. This foundation is further expanded in collaborative work with Dix et al. (2025), which introduces a benchmark of adversarial prompts designed to stress-test safety guardrails in leading LLMs and leverages a combination of human and AI evaluation to streamline model assessment. These studies underscore the importance of evaluating models not only under ideal conditions but also in adversarial and culturally sensitive contexts, directly informing the methodology of the current research.

Together, these studies support the consensus that LLMs offer remarkable generative capabilities, but their real-world deployment requires careful consideration of fairness, cultural sensitivity, ethical risk, and safety. This pilot study builds on existing research by leveraging human evaluators to assess the ability of state-of-the-art multilingual LLMs to faithfully localise nuanced language, such as idioms and cultural holidays, from English to 24 dialects of 20 languages in a realistic business setting.

## 4. Methodology

### 4.1 Objective

This study investigates how well publicly available, leading large language models (LLMs) perform the task of translating and localising culturally nuanced language. The focus is on real-world use cases—specifically, scenarios in which marketing professionals with limited LLM expertise might rely on model outputs to localise copy from English to other languages. Marketing content often includes humour, cultural references, and idiomatic expressions, making it an ideal test case for evaluating multilingual performance and cross-cultural generalisation.

### 4.2 Materials

Three anonymised marketing emails were adapted from real commercial campaigns [Appendix 1]. These included seasonal and culturally specific references (e.g., Valentine's Day and Singles Day) and products related to food and body image. Idiomatic language, such as humour and puns, was deliberately incorporated to test LLMs' ability to preserve tone and intent when translating culturally nuanced language.

## 4.3 Model Selection

To simulate realistic usage, we selected three widely known, freely available LLMs accessible to the general public at the time of the experiment. While models were anonymised in the study, the goal was not benchmarking per se, but a "state-of-use" snapshot of how LLMs perform in a practical, high-stakes task.

## 4.4 Participants

Twenty-two participants were recruited through convenience sampling. All had prior experience with LLM-related projects and were fluent in English and at least one other language. Some participants were multilingual, enabling comparison across multiple target languages. Each participant evaluated translations into their language(s) of fluency.

Language resource availability was measured in this study by global speaker population (large = >200m, medium = 100m–200m, small = <100m) and geographic distribution (global = multiple regions, regional = limited geographic region).

Table 1: *The study included 24 dialects across 20 languages, ranging in size (global speaker population) and geographic distribution (regional or global)*

| Language | Participant Region(s) | Size | Global or Regional |
|---|---|---|---|
| Farsi | Iran | small | regional |
| Finnish | Finland | small | regional |
| French | France, USA | medium | global |
| German | Germany, USA | medium | global |
| Gujarati | India | small | regional |
| Hebrew | Israel | small | regional |
| Hindi | India | large | global |
| Igbo | USA | small | regional |
| Italian | Italy | small | regional |
| Japanese | Japan | medium | regional |
| Korean | South Korea | small | regional |
| Malayalam | India | small | regional |
| Mandarin | China | large | global |
| Portuguese | Brazil | medium | global |
| Russian | Russia | medium | global |
| Spanish | Dominican Republic, Mexico, Spain, USA | large | global |
| Swedish | Sweden | small | regional |
| Tamil | India | small | regional |
| Urdu | Pakistan | medium | regional |

## 4.5 Procedure

Participants were given a standard prompt structure: "Translate the following email for use in [language] in [country/region]. [Email text]"

Each participant received three anonymised outputs (one per model) for each email. They were asked to evaluate the outputs on four criteria, as follows:

- Content fidelity
- Tone fidelity
- Cultural and audience appropriateness
- Overall localisation quality

Table 2: *Number of evaluations per language, variation based on convenience sampling for this pilot*

| Language | Number of evaluations |
|---|---|
| Farsi | 3 |
| Finnish | 3 |
| French | 9 |
| German | 12 |
| Gujarati | 3 |
| Hebrew | 3 |
| Hindi | 6 |
| Igbo | 3 |
| Italian | 3 |
| Japanese | 3 |
| Korean | 3 |
| Malayalam | 3 |
| Mandarin | 3 |
| Portuguese | 3 |
| Russian | 3 |
| Spanish | 15 |
| Swedish | 3 |
| Tamil | 3 |
| Urdu | 3 |
| Grand Total | 87 |

Each criterion was rated using a four-level scale, with "serious failures exist" on the low end and "very good or nearly perfect" on the high end:

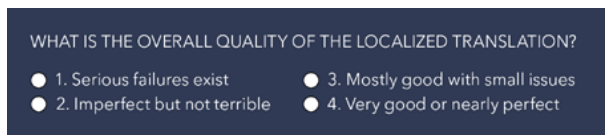

Figure 1: *Screenshot of a sample question showing the user interface in Appen's AI Data Platform where participants evaluated each output across the four-level scale.*

Participants then selected the best translation and indicated whether it was ready for use or required further editing. If edits were necessary, participants submitted improved versions and optionally provided qualitative comments on the set of responses.

This methodology enables both quantitative comparison and qualitative insight into model performance across languages, content types, and localisation challenges.

## 5. Results

### 5.1 Overall Model Performance

The three LLMs tested demonstrated comparable performance; therefore, the data is presented anonymously to highlight cross-language patterns rather than comparative benchmarking. Across the entire dataset, localisation quality varied significantly by language, even for identical inputs processed by the same model.

Table 3 shows the average localisation scores, normalised to percentages based on the four-level evaluation criteria described above, grouped by language.

### 5.2 Cross-Linguistic Patterns

*Language Family*
Languages more closely related to English generally achieved higher scores, though this relationship was inconsistent. Germanic languages like German (72.69%) and Swedish (77.78%) performed well, supporting the hypothesis that lexical overlap with English contributes to translation quality. However, Romance languages showed mixed results: French (76.85%) and Spanish (60.37%) performed strongly overall, while Portuguese (50.93%) and Italian (56.48%) lagged.

Indo-Aryan languages showed strong overall performance, particularly Hindi (78.70%) and Gujarati (71.30%), with Urdu (64.81%) scoring moderately lower. These results are consistent with prior research suggesting that LLMs perform more reliably in high-resource Indian languages like Hindi than in lower-resource regional languages like Urdu (Mujadia et al., 2023). Interestingly, Farsi (65.74%), an Indo-Iranian language, achieved moderate scores despite known limitations in training data and significant linguistic distance from English (Abaskohi, 2024).

*Orthography and Morphology*
While many high-performing languages used alphabetic systems (e.g., French, German, Spanish), others with alphabetic scripts (e.g., Portuguese, 50.93%; Igbo, 38.89%) performed poorly. Languages with syllabary scripts, Japanese (85.19%) and Korean (86.11%), performed exceptionally well. Among orthographies, the logographic script Mandarin (47.22%) demonstrated the lowest performance.

Table 3: *Average Scores by Language*

| Language | Continent | Language Family | Morphology | Orthography | Size | Distribution | Audience | Meaning | Tone | Overall Quality |
|---|---|---|---|---|---|---|---|---|---|---|
| Farsi | India | Indo-Aryan | Fusional | Abjad | small | regional | 70.37% | 55.56% | 77.78% | 59.26% |
| Finnish | Europe | Uralic | Agglutinative | Alphabetic | small | regional | 74.07% | 59.26% | 74.07% | 59.26% |
| French | Europe | Romance | Fusional | Alphabetic | medium | global | 79.01% | 74.07% | 76.54% | 77.78% |
| German | Europe | Germanic | Fusional | Alphabetic | medium | global | 75.93% | 67.59% | 77.78% | 69.44% |
| Gujarati | India | Indo-Aryan | Agglutinative | Abugida | small | regional | 66.67% | 77.78% | 66.67% | 74.07% |
| Hebrew | Europe | Semitic | Isolating | Abjad | small | regional | 66.67% | 59.26% | 62.96% | 66.67% |
| Hindi | India | Indo-Aryan | Fusional | Abugida | large | global | 79.63% | 79.63% | 75.93% | 79.63% |
| Igbo | Africa | Niger-Congo | Isolating | Alphabetic | small | regional | 40.74% | 37.04% | 37.04% | 40.74% |
| Italian | Europe | Romance | Fusional | Alphabetic | small | regional | 48.15% | 51.85% | 70.37% | 55.56% |
| Japanese | Asia | Japanese | Agglutinative | Syllabary | medium | regional | 92.59% | 77.78% | 96.30% | 74.07% |
| Korean | Asia | Korean | Agglutinative | Syllabary | small | regional | 92.59% | 85.19% | 88.89% | 77.78% |
| Malayalam | India | Dravidian | Agglutinative | Abugida | small | regional | 74.07% | 74.07% | 74.07% | 51.85% |
| Mandarin | Asia | Chinese | Isolating | Logographic | large | global | 48.15% | 44.44% | 51.85% | 44.44% |
| Portuguese | Europe | Romance | Fusional | Alphabetic | medium | global | 55.56% | 44.44% | 55.56% | 48.15% |
| Russian | Europe | Slavic | Fusional | Alphabetic | medium | global | 66.67% | 66.67% | 66.67% | 77.78% |
| Spanish | Europe | Romance | Fusional | Alphabetic | large | global | 61.48% | 60.74% | 65.93% | 53.33% |
| Swedish | Europe | Germanic | Fusional | Alphabetic | small | regional | 85.19% | 70.37% | 81.48% | 74.07% |
| Tamil | India | Dravidian | Agglutinative | Abugida | small | regional | 77.78% | 74.07% | 77.78% | 66.67% |
| Urdu | India | Indo-Aryan | Fusional | Abjad | medium | regional | 62.96% | 66.67% | 62.96% | 66.67% |
| Grand Total | | | | | | | 69.99% | 65.52% | 71.26% | 64.62% |

Similarly, morphology did not demonstrate a direct relationship with translation quality. Agglutinative languages like Japanese (85.19%), Korean (86.11%), Tamil (74.07%), and Malayalam (68.52%) were overrepresented among top performers. Fusional languages such as French, German, and Hindi also performed above the dataset's overall average (67.85%). In contrast, isolating languages like Igbo (38.89%) and Mandarin (47.22%) had the lowest scores.

*Resource Availability*
Languages with large speaker populations and global distribution, such as Spanish (60.37%) and Hindi (78.70%), generally achieved higher scores. However, performance among medium-sized and regionally limited languages was mixed: Swedish (77.78%) performed above average, while Portuguese (50.93%), despite its global use, underperformed. Small, regional languages, such as Igbo (38.89%), showed the weakest results overall. Mandarin (47.22%) further complicates this pattern, as it combines a large global speaker base with lower performance, likely due to tokenization challenges and script complexity.

Notably, several smaller, regional languages exceeded the dataset's overall average (67.85%), challenging the assumption that resource availability—defined by speaker population size and global distribution—is a reliable predictor of LLM translation quality. Table 4 explores these outliers in greater detail.

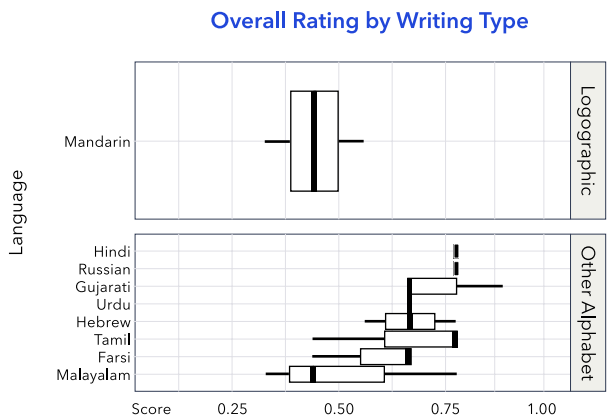

Figures 2 & 3: *Distribution of Overall Rating by Language and Writing System.*

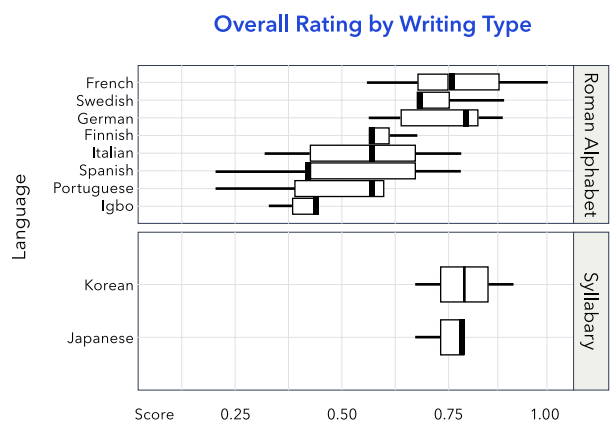

Table 4: *Positive Outliers Based on Language Size*

| Language | Average Across All Criteria | Interpretation |
|---|---|---|
| Korean | 86.11% | Strong model performance is likely supported by the tokenization advantages of its syllabary orthography and the accessibility of high-quality training data. |
| Swedish | 77.78% | Lexical overlap with English and a strong digital footprint are the most likely explanations for high translation quality, despite a small speaker base and regional use. |
| Gujarati | 71.30% | Small but globally dispersed, Gujarati's strong performance suggests targeted investment by model builders in Indian language datasets. |
| Malayalam | 68.52% | Similar to Gujarati, these results indicate focused investment from model builders to optimise for regional Indian languages. |

# 6. Discussion

This study demonstrates that while large language models (LLMs) generate translations across a wide range of languages with minimal grammatical errors, overall localisation quality remains inconsistent. All languages required human correction to achieve natural-sounding translations, faithful to the tone and meaning of the original English. Resource availability, as measured by speaker population and global distribution, and orthography emerged as the greatest predictors of localisation quality.

## 6.1 Idiomatic and Figurative Language as Persistent Challenges

Evaluators consistently reported that while most LLM translations were grammatically correct, they often sounded unnatural or overly literal. Idiomatic expressions and puns were the most consistent sources of translation errors across languages. English wordplay such as "Will you brie mine?", "cat's meow," and "feline good" was systematically mistranslated, resulting in awkward or confusing outputs that lost the intended humour or persuasiveness of the original.

The severity of these mistranslations varied across languages and was especially egregious in languages with significant structural and cultural distance from English, such as Mandarin, where direct translation of figurative expressions more frequently resulted in stiff or confusing outputs. Table 5 presents four examples of mistranslations across languages as well as the evaluator's feedback.

In contrast, successful translations adapted idioms creatively, using culturally relevant alternatives rather than literal translations. For example, one LLM translated "Singles Day" to "Pepero Day" for Korean, referencing a local holiday popular with single people. Similarly, a Swedish evaluator praised the use of "spinna av glädje" (purr with joy) to translate "feline good," demonstrating effective localisation. These examples underscore the need for models not only to translate words but to reinterpret meaning in culturally relevant ways.

## 6.2 Data Availability and Orthography Drive Performance

Model performance correlated most strongly with two factors: the availability of high-quality training data and the compatibility of the writing system with subword tokenisation methods. Linguistic proximity to English, by contrast, was not a reliable predictor of localisation quality.

Table 5: *Select Mistranslation Examples and Corresponding Evaluator Feedback*

| Language | Evaluator Feedback |
|---|---|
| Finnish | I changed the translation "keepsake tin" to "säilytettävään rasiaan" (instead of "muistorasiaan") because the word "muisto" can sometimes have an impression of something really gone, like forever, and it's not the best option here. |
| German | The models translated "cat's meow" as "katzenjammer" which has a nega-tive/sad connotation in German, and I am very sure that a shop or company would definitely not want to use this to advertise for something. |
| Italian | English and Italian are quite different languages, so puns have to be rewritten entirely. In this case, models #1 and #2 translated "will you brie mine?" as "Do you want to be my cheese?" which, albeit amusing, isn't probably appropriate. Model #3 went for a direct "Will you be my Brie?" which is frankly quite confusing. An alternative may be "Vuoi essere la mia fontina d'ispirazione?", i.e. "Will you be my [little inspiration source | fontina]?" Probably not the best wordplay, but beats calling your significant other a slice of cheese... |
| Mandarin | "Single's Day" should be translated as "单身日" instead of "光棍节" because "光棍" is a slang term referring to single men, which is not polite. |

- Korean and Japanese achieved the highest overall scores despite minimal linguistic overlap with English. Both languages benefit from simple syllabary orthographies and extensive digital footprints, making them well-represented in multilingual training datasets.
- Regional Indian languages (Gujarati, Tamil, Malayalam) outperformed expectations for "small, regional" languages. This suggests that targeted data inclusion by developers can elevate model performance even in languages with limited global reach.
- Languages with isolating or logographic scripts (Igbo, Mandarin) performed lowest, highlighting persistent tokenisation difficulties for these writing systems and, in Igbo's case, limited data availability (Hu, et al., 2024; Xue et al., 2021).

The interaction between script type and model architecture is particularly important. Japanese and Korean's success likely stems from their limited character sets, which simplify tokenisation. In contrast, logographic scripts like Mandarin and abjad scripts such as Farsi and Urdu face segmentation challenges that lead to weaker localisation results.

Morphological complexity alone did not predict localisation quality. Both fusional and agglutinative languages demonstrated strong performance when sufficient data was available. Instead, the data suggests that localisation success depends on two interacting factors: access to NLP training data and the degree to which a writing system aligns with model tokenisation strategies.

**6.3 Human Revision Remains Essential**

Despite the generally high grammatical accuracy of LLM translations, human revision consistently improved localisation quality across all languages. Evaluators frequently adjusted word choices, refined tone, and substituted culturally resonant alternatives for mistranslated idioms, puns, and cultural references. Even technically correct translations often sounded awkward, overly formal, or culturally mismatched without human intervention.

A recurring challenge was deciding when to preserve English terms versus adapting them for local audiences. Reviewers noted that untranslated English words sometimes enhanced authenticity in digital or product names but often felt out of place when used in other contexts. These findings suggest that current leading multilingual LLMs cannot yet be relied upon to consistently produce culturally aligned translations without expert human input.

# 7. Conclusion

This pilot demonstrates that while multilingual LLMs have achieved impressive levels of grammatical accuracy across diverse languages, they fall short when translating culturally nuanced language. Idioms, puns, and wordplay frequently resulted in literal, awkward, or contextually mismatched translations. These errors appeared across both high- and low-resource languages, challenging the notion that increased access to data equates to effective localisation. Human revision remained essential for achieving natural, engaging translations, particularly in marketing contexts where cultural resonance is pivotal to successful communication.

These findings underscore the opportunity for further investigation and potential for targeted efforts to improve multilingual LLM localisation capabilities, including enhanced training data diversity and deeper integration of cultural knowledge, particularly for applications requiring cultural sensitivity such as global marketing and customer communication.

# 8. Limitations

*Positivity of Feedback*
Evaluator comments rarely used strongly negative terms like "inaccurate" or "inappropriate," which may indicate either generally acceptable translation quality or a bias toward constructive or positive framing. This could result in underreporting of more severe translation errors.

*Language and Regional Representation*

The dataset heavily represented Romance, Indo-Aryan, and Germanic languages families with greater linguistic similarity to English. Most feedback came from evaluators in Europe, the Americas, and India, potentially limiting the generalisability of findings to more distant languages and regions.

*Scale of the Dataset*

With 87 evaluated units and fewer than three evaluators per language, this research represents a small pilot study. The focus on a single domain—e-commerce email marketing—further limits the applicability of findings to other content types or industries. Future work will expand this pilot to increase the robustness of the analysis.

*Evaluator Expertise*

While all evaluators were experienced in LLM evaluation tasks, their linguistic expertise and translation experience were not systematically controlled. This variability may have influenced both the depth and focus of feedback across languages.

*LLM Model Diversity*

The study anonymised model identities to reflect real-world user experience rather than comparative benchmarking. While appropriate for this research aim, this choice limits insights into specific strengths or weaknesses across different model architectures or providers.

**Appendix 1**

**Email 1**
Company: Sheffield's – a gourmet market in NYC
Subject: Will you brie mine?

Valentine's Day is almost here, and we've got the sweetest gift ideas for pickup or delivery throughout NYC.

Cheese Tasting Gift Boxes
This cheese lover's dream is thoughtfully assembled by our expert cheesemongers. It all comes beautifully packaged in a keepsake tin, tied with a satin ribbon. Personalize it with a custom note on Sheffield's stationary.

Valentine's Charcuterie Boards
Artfully displayed with the perfect accompaniments of fresh & dried fruit, nuts, honey, fig jam, espresso brownies, dark chocolate-covered strawberries, Valentine's candies, edible flowers and sliced baguette.

[order here]
We still have a limited number of handmade, chocolate-covered strawberries and
floral arrangements available for pre-order! Give us a call today or stop by the shop
before they're gone.

Wishing you a sweet Valentine's Day!
Sheffield's – Park Slope
Brooklyn, NY

**Email 2**
Company: Terra – an eco-friendly deodorant brand
Subject: This scent will transform your love life 😘✨

Swipe right this Single's Day
Use code: SINGLESDAY
[shop deodorant]

Hey [NAME]
This Singles' Day, give your deodorant game an irresistible upgrade – pair our newest reusable case design with a fragrance that's sure to make memories. Durable, stylish, compact, and zero waste.

Whether you're keeping yourself fresh for your partner, or looking to impress someone else, our new scents will leave a lasting impression.

MIX & MATCH OUR BEST-SELLING COMBOS
Lavender case x Tropical Paradise scent
Turquoise case x Orange Creamsicle scent

WHY TERRA?
Aluminum & paraben free. Zero-waste refills. 24-hour odor protection. All that in a case you'll be excited to reuse.

Terra Cosmetics
London N1C 4AB, United Kingdom

**Email 3**
Company: Muggable – an American novelty mug company
Subject: This Collection Has Us Feline Good

CAT'S MEOW
Our newest collection is the cat's pajamas, wait no – it's the cat's Mugs, Tumblers, Koozies, and Coasters!
[Shop Meow]
Rep your favorite feline at the office, on the go, and on your next Zoom call. Wait.Who are we kidding? They're already in all your Zoom calls.

©2012 Muggable Inc. All Rights Reserved.
Los Angeles, C A, 900 13, USA